\documentclass[letterpaper, 10 pt, conference]{ieeeconf}
\IEEEoverridecommandlockouts
\usepackage{cite}
\usepackage{graphics} % for pdf, bitmapped graphics files
\usepackage{epsfig} % for postscript graphics files
\usepackage{mathptmx} % assumes new font selection scheme installed
\usepackage{times} % assumes new font selection scheme installed
\usepackage{amsmath} % assumes amsmath package installed
\usepackage{amssymb}  % assumes amsmath package installed
\usepackage{graphicx}
\usepackage[table]{xcolor}
\definecolor{turquoise}{RGB}{0,160,180}
\usepackage{booktabs}
\usepackage{array}
\usepackage{tabularx}
\usepackage{ragged2e}
\usepackage{placeins}

\makeatletter
\let\NAT@parse\undefined
\makeatother
\usepackage[hidelinks]{hyperref}

\definecolor{tabheader}{RGB}{250,250,239}
\definecolor{tabours}{RGB}{235,248,226}

\DeclareSymbolFont{cmsymbols}{OMS}{cmsy}{m}{n}
\DeclareSymbolFontAlphabet{\mathcalcm}{cmsymbols}

\newcommand{\tabicon}[1]{%
    \raisebox{-0.12\height}{%
        \includegraphics[height=8.2pt]{#1}}%
}

\newcommand{\mono}{\tabicon{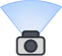}}
\newcommand{\surround}{\tabicon{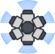}}
\newcommand{\pano}{\tabicon{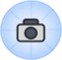}}
\newcommand{\videos}{\tabicon{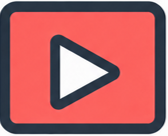}}
\newcommand{\car}{\tabicon{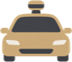}}
\newcommand{\quadrobot}{\tabicon{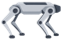}}
\newcommand{\human}{\tabicon{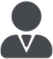}}
\newcommand{\vfm}{\tabicon{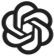}}
\newcommand{\synthetic}{\tabicon{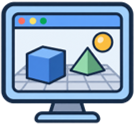}}
\newcommand{\wheels}{\tabicon{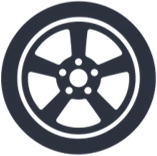}}
\newcommand{\gait}{\tabicon{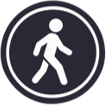}}
\newcommand{\stationary}{\tabicon{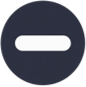}}
\newcommand{\yesmark}{\tabicon{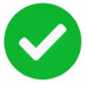}}
\newcommand{\nomark}{\tabicon{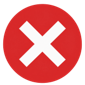}}

\newcommand{\classhead}[3]{%
    \raisebox{0pt}[1.30cm][0pt]{%
        \begin{tabular}[b]{@{}cc@{}}
            \rotatebox[origin=bl]{90}{#1}
            &
            \rotatebox[origin=bl]{90}{(#2)}
            \\[0pt]
            \multicolumn{2}{c}{%
                \textcolor{#3}{\rule{1.7ex}{1.7ex}}%
            }
        \end{tabular}%
    }%
}

\newcolumntype{L}[1]{%
    >{\hsize=#1\hsize\linewidth=\hsize\raggedright\arraybackslash}X}
\newcolumntype{Y}[1]{%
    >{\hsize=#1\hsize\linewidth=\hsize\centering\arraybackslash}X}

\definecolor{floorc}{RGB}{165,165,165}
\definecolor{wallc}{RGB}{125,90,55}
\definecolor{ceilingc}{RGB}{65,165,75}
\definecolor{humanc}{RGB}{255,65,70}
\definecolor{doorc}{RGB}{80,125,195}
\definecolor{glassc}{RGB}{205,165,65}
\definecolor{hydrantc}{RGB}{155,90,185}
\definecolor{otherc}{RGB}{55,175,175}
\definecolor{oursgray}{RGB}{232,232,232}

\newcolumntype{C}{>{\centering\arraybackslash}X}

\title{\LARGE \bf
CrowdOcc: Monocular Semantic Scene Completion for Quadruped Robots in Crowded Indoor Environments
}

\author{
\begin{tabular}{c}
Feiyang Chen$^{1}$,
Jincheng Hu$^{1*}$,
Yiduo Chen$^{1}$,
Jihao Li$^{2}$ \\[2pt]
Yue Liang$^{1}$,
Bingzhao Gao$^{1}$,
Yanjun Huang$^{1}$,
Yuanjian Zhang$^{1}$
\end{tabular}%
\thanks{$^{1}$Tongji University, Shanghai, China.}
\thanks{$^{2}$Loughborough University, Loughborough, United Kingdom.}
\thanks{$^{*}$Corresponding author.
\href{mailto:Jincheng_hu@tongji.edu.cn}{Jincheng\_hu@tongji.edu.cn}}
}

\begin{document}

\maketitle
\thispagestyle{empty}
\pagestyle{empty}

%%%%%%%%%%%%%%%%%%%%%%%%%%%%%%%%%%%%%%%%%%%%%%%%%%%%%%%%%%%%%%%%%%%%%%%%%%
\begin{abstract}

Monocular semantic scene completion (SSC) for quadruped robots remains underexplored in real crowded indoor environments, where human-scene occlusion disrupts static geometry and human occupancy predictions are often incomplete or spatially misplaced. We present CrowdOcc, an RGB-D dataset and monocular SSC framework for this setting. CrowdOcc contains 25.1K frames from 11 indoor scenes, with semantic occupancy annotations constructed through static-dynamic decoupling. Our framework combines: (i) Normal-Guided Scene Geometry Fusion (NGSGF) to complement depth-aware lifting with surface-normal cues for occlusion-robust geometry; and (ii) Human-Centric Sparse Interaction (HCSI) to selectively model human-human and local human-scene relations in 3D. Our method achieves state-of-the-art SSC performance on CrowdOcc's scene-disjoint test set, reaching 15.80 IoU, 11.40 mIoU, and 46.23 Human IoU, demonstrating generalization to unseen indoor scenes.

\end{abstract}

\vspace{6mm}
%%%%%%%%%%%%%%%%%%%%%%%%%%%%%%%%%%%%%%%%%%%%%%%%%%%%%%%%%%%%%%%%%%%%%%%%%%%%%%%%
\section{INTRODUCTION}

Semantic Scene Completion (SSC) predicts dense 3D occupancy and semantics from partial observations~\cite{song2017sscnet,wang2019forknet,li2019ddrnet,li2020palnet,li2020aicnet,chen2020sketchssc,cai2021sisnet}. Monocular SSC further reduces the sensing requirement to a single RGB image, making it attractive for onboard perception on quadruped robots~\cite{cao2022monoscene}. However, most existing SSC datasets are built around static indoor scenes or autonomous-driving settings, where the sensing viewpoint, scene dynamics, and occupancy distribution differ substantially from those encountered by quadruped robots in crowded indoor environments~\cite{yu2024iso,behley2019semantickitti,tian2023occ3d,shi2026oneocc}. In particular, existing datasets provide limited coverage of low-viewpoint observations with dense human presence and dynamic human occupancy. As summarized in Table~\ref{tab:dataset_comparison}, this leaves real crowded indoor SSC for quadruped robots largely underexplored.

Crowded indoor scenes introduce two difficulties that are not well addressed by conventional monocular SSC pipelines. First, humans frequently occlude large portions of walls, floors, doors, and other static structures. The resulting incomplete observations weaken depth-based feature lifting and make the recovery of scene geometry less reliable~\cite{yao2025depthssc,lee2025soap}. Second, human occupancy remains difficult to recover reliably in crowded scenes. Partial observations and foreground-background ambiguity frequently lead to incomplete human geometry and erroneous spatial placement. These issues become more pronounced when the test environment differs from the training scenes, where scene-specific geometric priors no longer transfer reliably.

\begin{figure}[!t]
    \centering
    \includegraphics[width=\columnwidth]{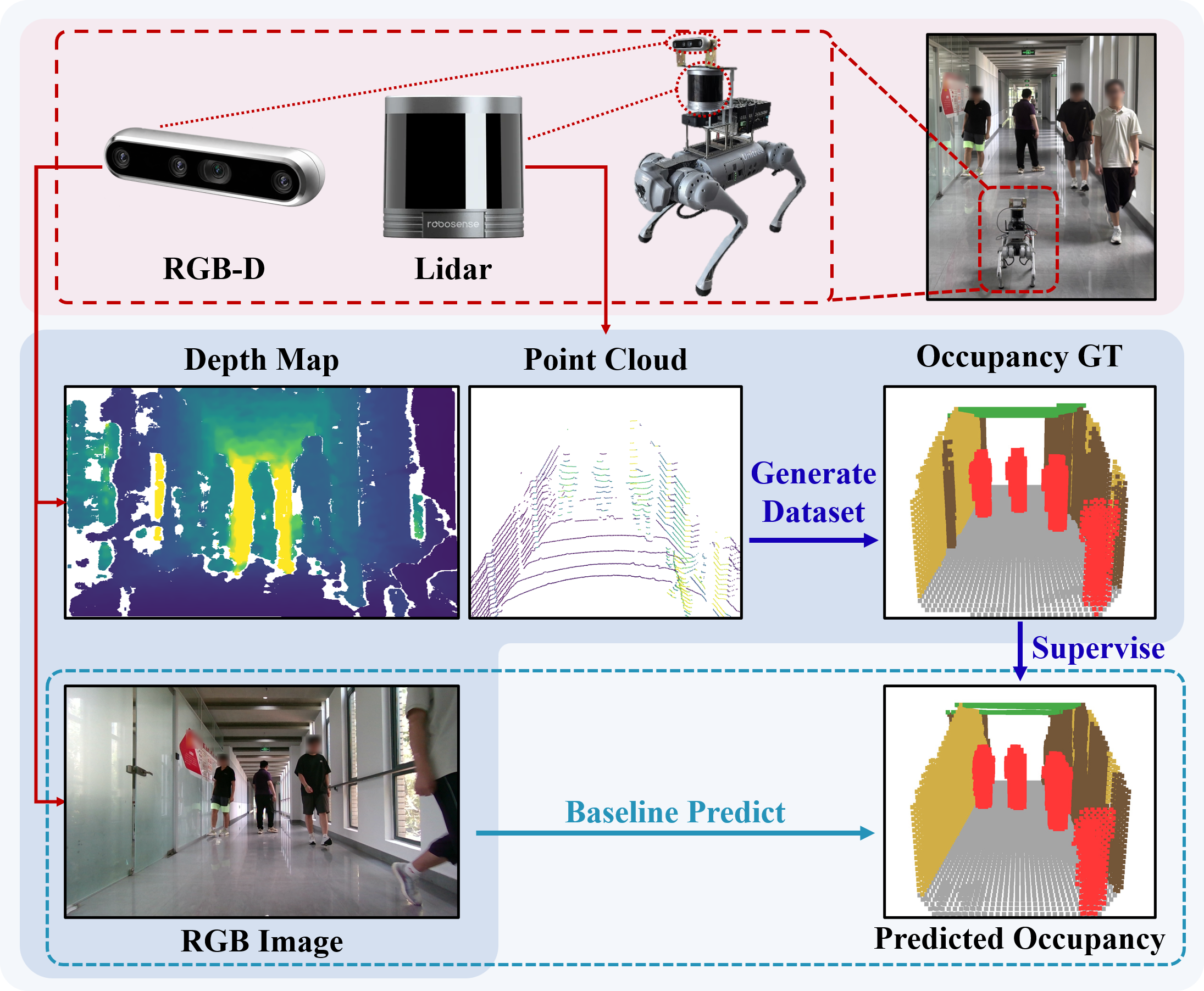}
    \vspace{-6mm}
    \caption{\textbf{CrowdOcc Overview.}
    Synchronized RGB-D, LiDAR, and IMU observations collected by a quadruped robot are used for the offline construction of semantic occupancy annotations. The resulting dataset provides RGB-D frames and semantic occupancy annotations for monocular SSC, where inference takes a single RGB image as input. The provided depth maps can optionally serve as auxiliary supervision during training. The \textcolor{red}{red} background denotes the sensing hardware, the \textcolor{blue}{blue} background indicates the CrowdOcc dataset construction pipeline, while the \textcolor{cyan}{turquoise} dashed box shows the model prediction pipeline.}
    \label{fig:crowdocc_overview}
    \vspace{-5mm}
\end{figure}

To address this difficulties, we introduce CrowdOcc, an RGB-D dataset for monocular SSC in real crowded indoor environments, together with a model tailored to scene geometry and human occupancy. CrowdOcc contains 25.1K frames collected from 11 independent indoor scenes using a quadruped robot. As illustrated in Fig.~\ref{fig:crowdocc_overview}, synchronized RGB-D, LiDAR, and IMU measurements are used for offline occupancy construction, while inference takes only a single RGB image. For scene geometry, we propose Normal-Guided Scene Geometry Fusion (NGSGF) module, which complements depth-based lifting with surface-normal cues and improves the representation of partially observed static structures. For human occupancy, we introduce Human-Centric Sparse Interaction (HCSI) module, which selects human-related voxel candidates and explicitly models human-human and local human-scene interactions in 3D. We evaluate cross-scene generalization on test scenes excluded from both training and validation. Our method achieves state-of-the-art SSC performance under this protocol.

\begin{table*}[!t]

% ============================================================
% Caption
% ============================================================
\refstepcounter{table}
\label{tab:dataset_comparison}

\begin{minipage}{\textwidth}
\normalsize
\setlength{\parindent}{0pt}
\justifying
\noindent
TABLE~\thetable:
Typical datasets for Semantic Scene Completion (SSC).
Abbreviations:
\mono~(Monocular View),
\surround~(Multi-camera Surround View),
\pano~(Panoramic View),
\videos~(Internet images/videos),
\car~(Autonomous Car),
\quadrobot~(Quadruped Robot),
\human~(Human),
\vfm~(Vision Foundation Model),
\synthetic~(Synthetic Data),
\wheels~(Wheels),
\gait~(Gait),
\stationary~(Stationary).
Plat. (Platform), Crowd (Crowded Scenario), Move. (Movement),
Env. (Environment), Anno. (Annotation),
C (Camera), D (Depth), L (LiDAR).
\end{minipage}

% -------- Caption 与表格间距：主要调这里 --------
\vspace{5pt}

% ============================================================
% Table
% ============================================================
{
\footnotesize

% 行距
\renewcommand{\arraystretch}{1.65}

% 普通列左右 padding
\setlength{\tabcolsep}{2pt}

\arrayrulecolor{black}

\begin{tabularx}{\textwidth}{
    @{}
    L{2.60}
    Y{0.65}
    Y{0.65}
    Y{0.65}
    Y{0.70}
    Y{0.85}
    Y{0.95}
    Y{1.45}
    Y{0.90}
    Y{0.85}
    Y{0.75}
    @{}
}

% ============================================================
% 最上方粗线
% ============================================================
\specialrule{1.20pt}{0pt}{0pt}

% ============================================================
% Header
% ============================================================
\rowcolor{tabheader}
\textbf{Dataset}
& \textbf{View}
& \textbf{Plat.}
& \textbf{Crowd}
& \textbf{Move.}
& \textbf{Env.}
& \textbf{Modality}
& \textbf{Volume Size}
& \textbf{Voxel (m)}
& \textbf{Frames}
& \textbf{Anno.} \\

% 两条很细的线
\specialrule{0.30pt}{0pt}{0.55pt}
\specialrule{0.30pt}{0pt}{0pt}

NYUv2~\cite{silberman2012nyuv2}
& \mono
& \videos
& \nomark
& \stationary
& Indoor
& C\&D
& $144{\times}240{\times}240$
& 0.02
& 1.4K
& \human \\

Occ-ScanNet~\cite{yu2024iso}
& \mono
& \videos
& \nomark
& \stationary
& Indoor
& C
& $36{\times}60{\times}60$
& 0.08
& 65.5K
& \human \\

EmbodiedOcc-ScanNet\textsuperscript{\ensuremath{\dagger}}~\cite{wu2025embodiedocc}
& \mono
& \videos
& \nomark
& \stationary
& Indoor
& C
& Variable
& 0.08
& 20.2K
& \human \\

SemanticKITTI~\cite{behley2019semantickitti}
& \mono
& \car
& \nomark
& \wheels
& Outdoor
& C\&L
& $32{\times}256{\times}256$
& 0.20
& 9K
& \human \\

SSCBench-KITTI-360~\cite{li2024sscbench,liao2023kitti360}
& \mono
& \car
& \nomark
& \wheels
& Outdoor
& C\&L
& $32{\times}256{\times}256$
& 0.20
& 12.9K
& \human \\

Occ3D-nuScenes~\cite{tian2023occ3d,caesar2020nuscenes}
& \surround
& \car
& \nomark
& \wheels
& Outdoor
& C\&L
& $16{\times}200{\times}200$
& 0.40
& 40K
& \human \\

WildOcc~\cite{zhai2024wildocc}
& \mono
& \car
& \nomark
& \wheels
& Outdoor
& C\&L
& $40{\times}100{\times}100$
& 0.20
& 10K
& \human \\

Human360Occ~\cite{shi2026oneocc}
& \pano
& \human
& \nomark
& \gait
& Outdoor
& C\&D
& $16{\times}128{\times}128$
& 0.20
& 8K
& \synthetic \\

QuadOcc~\cite{shi2026oneocc}
& \pano
& \quadrobot
& \nomark
& \gait
& Outdoor
& C\&L
& $8{\times}64{\times}64$
& 0.40
& 24K
& \vfm\,\human \\

% ============================================================
% 第三条线：中等粗度
% ============================================================
\specialrule{0.65pt}{0pt}{0pt}

% ============================================================
% Ours
% ============================================================
\rowcolor{tabours}
\textbf{CrowdOcc (ours)}
& \mono
& \quadrobot
& \yesmark
& \gait
& Indoor
& C\&D
& $50{\times}60{\times}96$
& 0.08
& 25.1K
& \vfm\,\human \\

% ============================================================
% 最下方粗线
% ============================================================
\specialrule{1.20pt}{0pt}{0pt}

\end{tabularx}
}

% ============================================================
% Table note
% ============================================================
\vspace{1.5pt}

\begin{minipage}{\textwidth}
\scriptsize
\setlength{\parindent}{0pt}
\justifying
\noindent
\textsuperscript{\ensuremath{\dagger}}EmbodiedOcc-ScanNet uses
$36{\times}60{\times}60$ local occupancy annotations, while the global
occupancy size varies with the spatial extent of each scene~\cite{wu2025embodiedocc}.
\end{minipage}

\vspace{-4mm}

\end{table*}

Our key contributions are:
\begin{itemize}
    \setlength{\itemsep}{2pt}
    \setlength{\parskip}{0pt}
    \setlength{\parsep}{0pt}
    \setlength{\topsep}{2pt}

    \item \textbf{Dataset.} We introduce CrowdOcc, an RGB-D dataset for monocular SSC in real crowded indoor environments from a quadruped viewpoint, comprising 25.1K frames from 11 independent scenes, 8 occupied semantic categories, and 0.08 m voxels on a \(50\times60\times96\) grid.

    \item \textbf{Method.} We propose NGSGF module and HCSI module to address the two main difficulties of crowded indoor SSC. NGSGF enhances static-scene geometry under human occlusion by coupling depth-based lifting with surface-normal cues, while HCSI refines human geometry and spatial placement through sparse human-centric 3D interaction.

    \item \textbf{Generalization.} Extensive experiments on scene-disjoint test scenes show strong cross-scene generalization, with our method achieving state-of-the-art SSC performance on unseen indoor environments.
\end{itemize}

\section{RELATED WORK}

\subsection{Monocular Semantic Scene Completion}

Monocular SSC recovers dense 3D occupancy and semantics from a single RGB image~\cite{yao2025depthssc,wang2025monomrn}. MonoScene~\cite{cao2022monoscene} established a representative image-to-voxel pipeline by lifting 2D features into a 3D volume and performing volumetric context reasoning. Subsequent methods improve either the lifting process or the resulting 3D representation~\cite{liu2024fullysparseocc}. VoxFormer~\cite{li2023voxformer} constructs depth-guided sparse voxel queries before dense completion, while NDC-Scene~\cite{yao2023ndcscene} performs feature projection in normalized device coordinates to reduce lifting ambiguity. ISO~\cite{yu2024iso} further incorporates monocular depth into indoor SSC through depth-aware feature projection and multi-scale fusion. Recent approaches explore instance-level context, scene-adaptive decoding, Gaussian representations, and additional geometric priors to improve monocular occupancy prediction~\cite{jiang2024symphonies,lee2025soap,jevtic2025scenedino,huang2024gaussianformer,huang2025gaussianformer2,qian2026splatssc,zhou2026gpocc}.
Despite these advances, depth remains the dominant geometric cue for image-to-voxel lifting. Under heavy occlusion, depth alone provides limited information about local surface orientation and structural continuity. Our NGSGF complements depth-based lifting with surface-normal cues to improve static-scene geometry in crowded environments.

\subsection{Human-Centric and Embodied Occupancy Perception}

Vision-based occupancy prediction has been extensively studied in autonomous-driving settings~\cite{huang2023tpvformer,zhang2023occformer,tang2024sparseocc}, with recent efforts extending it to embodied platforms. EmbodiedOcc~\cite{wu2025embodiedocc} studies online occupancy prediction for vision-based robotic scene understanding, while EmbodiedOcc++~\cite{wang2025embodiedoccplus} further improves geometric modeling for robotic perception. OneOcc~\cite{shi2026oneocc} considers SSC for legged robots using panoramic vision and develops platform-specific designs for gait-induced motion and omnidirectional sensing. Human-aware occupancy has also been explored for modeling human-scene spatial relations and dynamic agents~\cite{liu2024humanoccupancy}.
Existing instance-aware SSC methods model instance and scene context~\cite{jiang2024symphonies,liu2025disc}. Our focus is targeted human-centric 3D reasoning to improve the completeness and spatial placement of human occupancy in crowded indoor scenes.

\begin{figure*}[!t]
    \centering
    \includegraphics[width=\textwidth]{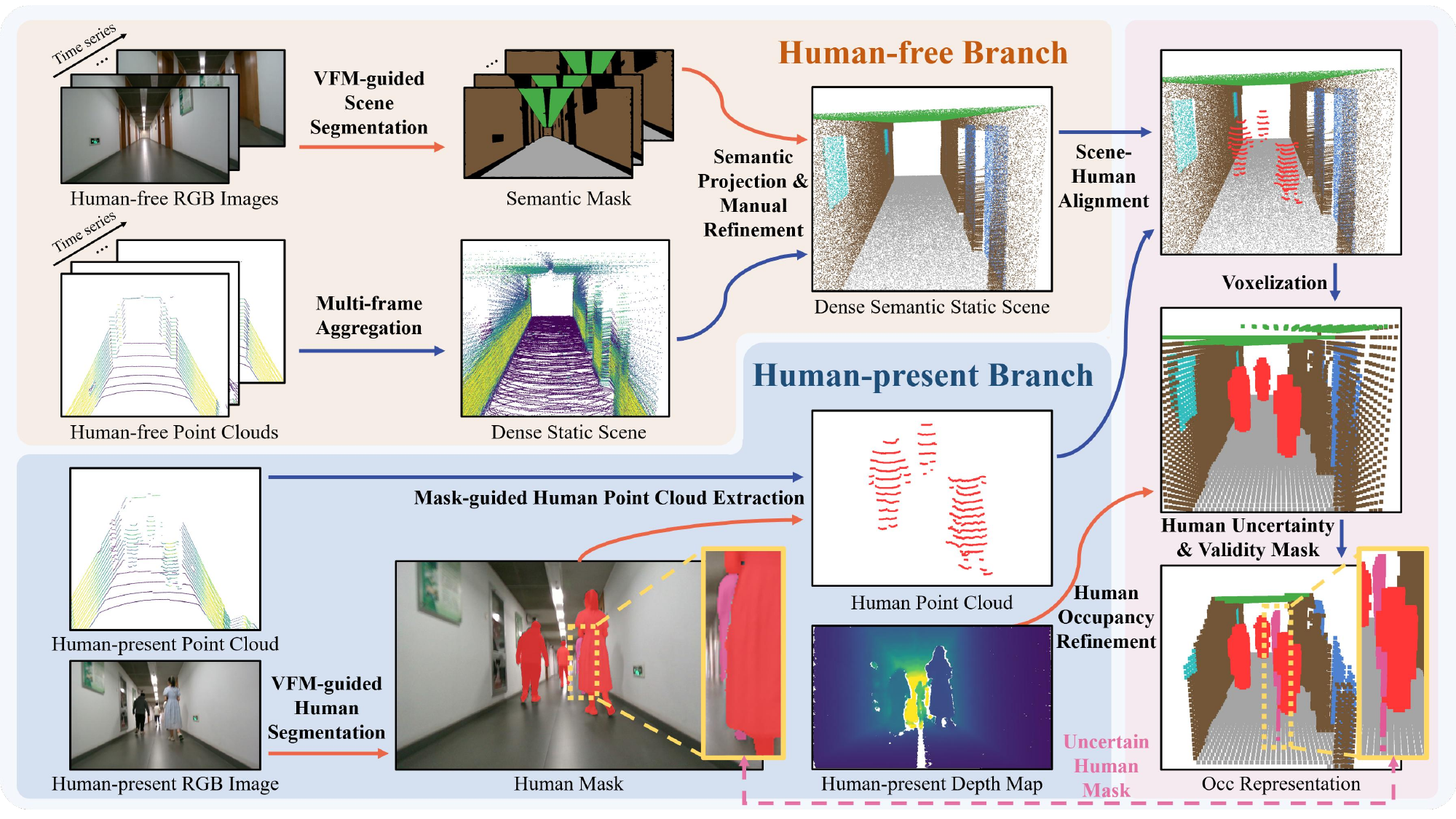}
    \vspace{-6mm}
    \caption{\textbf{Pipeline of our CrowdOcc dataset generation.}
    We adopt a dual-branch pipeline for human-free and human-present sequences. The human-free branch reconstructs a dense semantic static scene, while the human-present branch extracts human observations and aligns them with the static scene. During voxelization, the synchronized current-frame depth map is used to refine human occupancy labels, followed by uncertainty and validity masking to produce the final semantic occupancy annotations.}
    \label{fig:dataset_pipeline}
    \vspace{-4mm}
\end{figure*}

\subsection{SSC Datasets and Occupancy Construction}

Existing SSC datasets cover a wide range of sensing platforms and environments~\cite{ma2024cam4docc}. NYUv2~\cite{silberman2012nyuv2} and Occ-ScanNet~\cite{yu2024iso,dai2017scannet} focus primarily on static indoor scenes, whereas SemanticKITTI~\cite{behley2019semantickitti}, SSCBench-KITTI-360~\cite{li2024sscbench,liao2023kitti360}, and Occ3D~\cite{tian2023occ3d,caesar2020nuscenes} are designed for autonomous-driving settings. WildOcc~\cite{zhai2024wildocc} extends occupancy prediction to off-road environments. More recent datasets consider embodied and panoramic perception: Human360Occ provides synthetic human-ego panoramic occupancy, while QuadOcc collects real panoramic data from a quadruped robot~\cite{shi2026oneocc}. As summarized in Table I, none of these datasets jointly provide real crowded indoor scenes, low-viewpoint quadruped observations, and dynamic human occupancy.
Dense occupancy labels are commonly constructed by aggregating multi-frame observations. SurroundOcc~\cite{wei2023surroundocc} and OpenOccupancy~\cite{wang2023openoccupancy} use temporal sensor accumulation to improve scene coverage, while Occ3D~\cite{tian2023occ3d} additionally models visibility when defining occupied, free, and unobserved regions. Such aggregation is well suited to static geometry but is less reliable for dynamic objects. PointSSC~\cite{yan2024pointssc} therefore separates static and dynamic components and registers dynamic observations across views. This assumption becomes problematic for humans, whose pose and shape vary continuously over time. CrowdOcc instead reconstructs the static scene from human-free sequences and derives human occupancy from synchronized current-frame observations, avoiding temporal accumulation of non-rigid human geometry.

\section{CROWDOCC DATASET}

CrowdOcc comprises 25,140 RGB-D frames captured at 11 indoor scenes using a quadruped robot. Training and validation share eight scenes, containing 16,961 and 4,265 frames, respectively. The test set contains 3,914 frames from three additional scenes disjoint from both training and validation, enabling cross-scene evaluation. Each frame is paired with a semantic occupancy grid of \(50\times60\times96\) at 0.08 m resolution. LiDAR and IMU measurements are used only for offline annotation construction, while inference takes a single RGB image. Since static structures can be reliably aggregated across observations whereas non-rigid humans cannot, we construct them separately. As shown in Fig.~\ref{fig:dataset_pipeline}, human-free sequences are aggregated to recover the static scene, while human occupancy is derived from synchronized current-frame observations and subsequently refined during voxelization.

\subsection{Static Semantic Scene Annotation}

We divide the static scene annotation pipeline into two stages. First, we perform multi-frame static scene reconstruction from human-free sequences. Then, we transfer image-level semantic annotations to the reconstructed 3D scene to obtain dense semantic labels.

\textbf{Multi-frame Static Scene Reconstruction.}
We use human-free observations to avoid introducing transient human geometry into the static background. Point clouds from different frames are registered into a unified coordinate system and aggregated to reconstruct a dense static scene. Compared with single-frame observations, multi-frame aggregation increases surface coverage and recovers static structures that are partially occluded or sparsely observed from individual viewpoints.

\textbf{Semantic Static Scene Annotation.}
Directly annotating dense 3D scenes is labor-intensive.
We employ Grounded SAM 2~\cite{liu2024groundingdino,ravi2025sam2}
to generate semantic masks from the corresponding human-free
RGB images, followed by calibrated image-to-3D projection for
semantic association. Since individual views may contain incomplete or noisy semantic observations, projected labels from multiple frames are integrated over the static scene and manually refined to obtain consistent 3D semantic annotations. By constructing geometry and semantics exclusively from human-free sequences, the static branch provides a stable and dense background representation for subsequent dynamic human occupancy generation.

\subsection{Dynamic Human Occupancy Construction}

Dynamic humans continuously change pose and shape, making direct multi-frame aggregation unsuitable. We therefore construct human occupancy from synchronized current-frame observations and align it with the reconstructed static scene before voxelization.

\textbf{Current-frame Human Extraction and Alignment.}
We first use the same vision foundation model
~\cite{liu2024groundingdino,ravi2025sam2}
to obtain human masks from the current RGB image. The synchronized point cloud is projected onto the image plane, and points within the human regions are extracted as current-frame human observations. These observations are then transformed into the coordinate system of the reconstructed static scene through scene-human alignment, avoiding temporal accumulation of non-rigid human geometry.

\textbf{Depth-guided Human Occupancy Refinement.}
Human points extracted from a single frame can be sparse and provide only coarse spatial localization. During voxelization, the synchronized depth map is therefore used to refine the placement of human occupancy labels within the aligned scene.

\textbf{Human Uncertainty and Validity Masking.}
A human visible in the RGB image may have insufficient 3D support in the synchronized point cloud because of severe occlusion or sparse point observations. We mark such regions as uncertain and exclude them from supervision rather than assigning unsupported occupancy or free-space labels. A separate validity mask removes geometrically invalid or ignored voxels according to the occupancy supervision protocol. These masks are applied after depth-guided refinement to produce the final semantic occupancy annotations.

\section{METHODOLOGY}

Our evaluation of representative monocular SSC baselines on CrowdOcc reveals two recurring failure modes. First, human-scene occlusion leads to discontinuous static geometry, particularly on wall and floor surfaces. We attribute this to the limited surface-orientation constraints of depth-based lifting and introduce Normal-Guided Scene Geometry Fusion (NGSGF) module to complement depth with surface-normal cues. Second, human occupancy is often incompletely reconstructed or misplaced relative to the surrounding scene, indicating ambiguity in human-scene spatial reasoning. This motivates Human-Centric Sparse Interaction (HCSI) module, which selects Top-\(K\) human candidates for human-human and local human-scene interaction.

\begin{figure*}[!t]
    \centering
    \includegraphics[width=\textwidth]{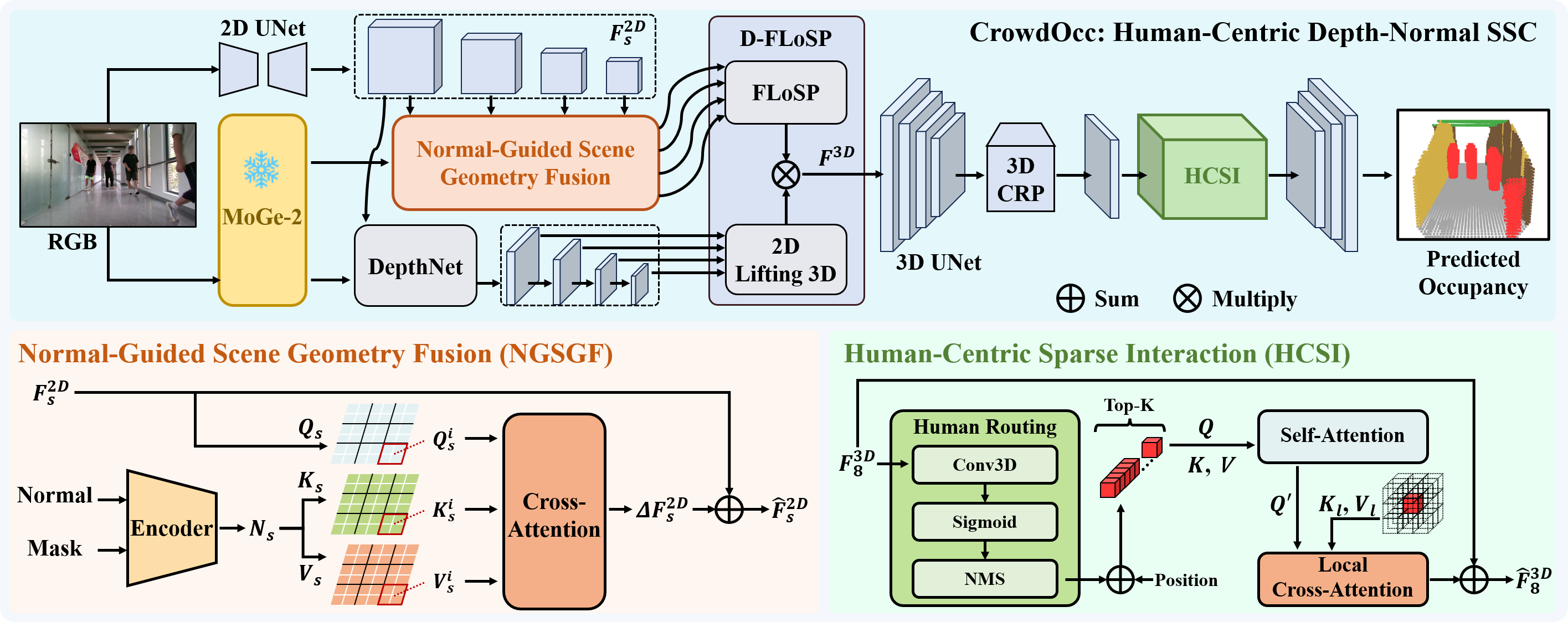}
    \vspace{-6mm}
    \caption{\textbf{Pipeline of our CrowdOcc framework.}
    Given a monocular RGB image, MoGe-2 estimates depth and surface normals from the same image. We use Normal-Guided Scene Geometry Fusion
    to enhance multi-scale 2D scene features before D-FLoSP lifting, and
    Human-Centric Sparse Interaction to selectively refine human-related 3D
    features. The enhanced representation is then decoded for semantic scene
    completion.}
    \label{fig:method}
    \vspace{-4mm}
\end{figure*}

Inspired by the depth-aware lifting paradigm of ISO~\cite{yu2024iso}, we formulate the overall architecture in Fig.~\ref{fig:method} around these two complementary designs.
Given an RGB image, a 2D UNet extracts multi-scale features \(F_s^{2D}\), while a frozen MoGe-2~\cite{wang2025moge2} estimates depth and surface normals. A trainable DepthNet combines the image features with MoGe-2 depth estimates to produce the multi-scale depth representation used by D-FLoSP. NGSGF enhances the 2D features before D-FLoSP lifting. The lifted representation is processed by the 3D encoder and CRP, followed by HCSI for sparse human-centric refinement and a 3D decoder for semantic occupancy prediction.

\subsection{Normal-Guided Scene Geometry Fusion}

Given the image feature $F_s^{2D}$ at scale $s$, we derive a
normal validity mask from the MoGe-2 outputs~\cite{wang2025moge2}
and numerical validity checks. The normal map and its validity
mask are jointly processed by a masked Normal Encoder to obtain
the geometry feature $N_s$. We then use $F_s^{2D}$ and $N_s$
to generate queries $Q_s$, keys $K_s$, and values $V_s$,
respectively:
\begin{equation}
Q_s = W_Q F_s^{2D}, \qquad
K_s = W_K N_s, \qquad
V_s = W_V N_s,
\label{eq:qkv_generation}
\end{equation}
where $W_Q$, $W_K$, and $W_V$ are learnable projections. To preserve local
geometric structures while avoiding the computational cost of global
cross-attention, we partition $Q_s$, $K_s$, and $V_s$ into local windows.
For the $i$-th window, the normal-guided interaction is formulated as
\begin{equation}
\Delta F_{s,i}^{2D}
=
\operatorname{Softmax}
\left(
\frac{
Q_s^i \left(K_s^i\right)^T
}{
\sqrt{d}
}
\right)
V_s^i,
\label{eq:normal_cross_attention}
\end{equation}
where $d$ denotes the feature dimension. The window-wise outputs are merged
to form the normal-guided residual $\Delta F_s^{2D}$, which is then fused with the original image representation:
\begin{equation}
\hat{F}_s^{2D}
=
F_s^{2D}
+
\gamma_s \Delta F_s^{2D},
\label{eq:normal_residual_fusion}
\end{equation}
where $\gamma_s$ is a learnable scaling coefficient.

\subsection{Human-Centric Sparse Interaction}

Given the input feature $F_8^{3D}$, a lightweight Human-Routing module first
predicts voxel-wise human responses. We apply local 3D NMS followed by Top-$K$
selection to obtain a sparse candidate set:
\begin{equation}
P_h
=
\operatorname{Sigmoid}
\left(
R(F_8^{3D})
\right),
\label{eq:human_probability}
\end{equation}
\begin{equation}
\Omega
=
\operatorname{TopK}
\left(
\operatorname{NMS}(P_h), K
\right),
\label{eq:human_candidates}
\end{equation}
where $R(\cdot)$ denotes the routing network and $\Omega$ contains the selected
candidate locations. The same prediction-based routing is used during both training and inference, while human occupancy labels are used only to supervise the routing scores. The selected candidate features interact through self-attention, which incorporates positional encodings of their 3D voxel locations:
\begin{equation}
F_h
=
\operatorname{SA}
\left(
F_8^{3D}[\Omega]
\right).
\label{eq:human_self_attention}
\end{equation}

For each selected candidate, we then perform cross-attention with its local 3D neighborhood sampled from the dense scene representation:
\begin{equation}
\Delta F_h
=
\operatorname{CA}
\left(
F_h,\,
\mathcal{N}(F_8^{3D})
\right),
\label{eq:human_cross_attention}
\end{equation}
where $\mathcal{N}(\cdot)$ denotes the local scene features centered at the candidate locations in $\Omega$. Finally, the
refined features are written back only to the selected locations:
\begin{equation}
\hat{F}_8^{3D}
=
F_8^{3D}
+
\gamma_h
\operatorname{Scatter}_{\Omega}
\left(
\Delta F_h
\right),
\label{eq:human_scatter}
\end{equation}
where $\gamma_h$ is a learnable residual scaling factor, and $\operatorname{Scatter}_{\Omega}(\cdot)$ writes the residuals to the locations indexed by $\Omega$ and assigns zero residuals elsewhere.

\subsection{Training Loss}

Following ISO~\cite{yu2024iso}, we retain its scene-completion and auxiliary depth objectives for training. The semantic occupancy annotations supervise scene completion, while the depth maps provided by CrowdOcc are used for the auxiliary depth objective. We denote these objectives together as
$\mathcalcm{L}_{base}$. Human occupancy labels are additionally used to supervise the Human-Routing module.
For human candidate prediction, binary cross-entropy loss
$\mathcalcm{L}_{BCE}$ and Dice loss
$\mathcalcm{L}_{Dice}$ are jointly applied to supervise the predicted
human probability $P_h$ with the human occupancy labels $Y_h$.
Invalid voxels are excluded from the supervision. The overall training
objective is formulated as:
{\setlength{\abovedisplayskip}{0pt}
 \setlength{\belowdisplayskip}{6pt}
 \setlength{\abovedisplayshortskip}{0pt}
 \setlength{\belowdisplayshortskip}{6pt}
 
\begin{equation}
\mathcalcm{L}
=
\mathcalcm{L}_{base}
+
\lambda_h
\left[
\mathcalcm{L}_{BCE}(P_h,Y_h)
+
\mathcalcm{L}_{Dice}(P_h,Y_h)
\right],
\label{eq:training_objective}
\end{equation}
}
where $\lambda_h$ is the balancing parameter for human-routing supervision.

\begin{table*}[!t]

% ============================================================
% Caption
% ============================================================
\refstepcounter{table}
\label{tab:baseline_comparison}

\begin{minipage}{\textwidth}
\footnotesize
\setlength{\parindent}{0pt}
\noindent
\justifying
TABLE~\thetable: Quantitative comparison on CrowdOcc.
Class frequencies, shown in parentheses, are computed over valid occupied voxels,
excluding empty and ignored/invalid voxels.
\textbf{Bold} indicates the best result and
\underline{underline} indicates the second-best result.
\end{minipage}

\vspace{5pt}

% ============================================================
% Table
% ============================================================
{
\scriptsize

\renewcommand{\arraystretch}{1.20}
\setlength{\tabcolsep}{1.4pt}
\arrayrulecolor{black}

\begin{tabularx}{\textwidth}{
    >{\raggedright\arraybackslash}p{2.00cm}
    >{\centering\arraybackslash}p{1.40cm}
    !{\vrule width 0.40pt}
    *{8}{C}
    !{\vrule width 0.40pt}
    >{\centering\arraybackslash}p{1.05cm}
    >{\centering\arraybackslash}p{1.05cm}
    !{\vrule width 0.40pt}
    >{\centering\arraybackslash}p{1.05cm}
    >{\centering\arraybackslash}p{1.05cm}
}

% ============================================================
% Top thick rule
% ============================================================
\specialrule{1.05pt}{0pt}{0pt}

\noalign{\vskip 1pt}

% ============================================================
% Header
% ============================================================
Method
&
Venue
&
\classhead{Floor}{28.97\%}{floorc}
&
\classhead{Wall}{35.19\%}{wallc}
&
\classhead{Ceiling}{13.93\%}{ceilingc}
&
\classhead{Human}{8.27\%}{humanc}
&
\classhead{Door}{6.53\%}{doorc}
&
\classhead{Glass}{6.30\%}{glassc}
&
\classhead{Hydrant}{0.55\%}{hydrantc}
&
\classhead{Other}{0.26\%}{otherc}
&
Precision$\uparrow$
&
Recall$\uparrow$
&
IoU$\uparrow$
&
mIoU$\uparrow$
\\

% ============================================================
% Header bottom rule
% ============================================================
\noalign{\vskip 1pt}
\noalign{\vskip 0.1pt}
\specialrule{0.40pt}{0pt}{0pt}
\noalign{\vskip 0.1pt}
\noalign{\vskip 1pt}

% ============================================================
% Baselines
% ============================================================

SplatSSC~\cite{qian2026splatssc}
& \mbox{AAAI 2026}
& 15.74
& 5.36
& 0.84
& 21.11
& 0.50
& 3.00
& 0.34
& 0.13
& 19.61
& 15.92
& 9.63
& 5.88
\\

GPOcc~\cite{zhou2026gpocc}
& \mbox{CVPR 2026}
& 16.99
& 6.53
& 0.93
& 32.51
& 1.80
& 2.42
& 1.67
& 0.67
& 20.21
& 18.18
& 10.58
& 7.94
\\

EmbodiedOcc~\cite{wu2025embodiedocc}
& \mbox{ICCV 2025}
& 19.54
& 6.76
& 1.40
& 30.89
& 1.10
& 2.13
& \underline{2.41}
& 0.10
& 19.39
& 21.60
& 11.38
& 8.04
\\

MonoScene~\cite{cao2022monoscene}
& \mbox{CVPR 2022}
& 19.17
& 9.53
& 1.52
& 43.09
& 1.57
& \underline{4.36}
& 1.69
& 0.00
& \textbf{21.67}
& 30.31
& 14.47
& 10.12
\\

NDC-Scene~\cite{yao2023ndcscene}
& \mbox{ICCV 2023}
& \underline{20.66}
& 9.68
& \textbf{2.09}
& \underline{44.12}
& \underline{3.34}
& 1.22
& 1.46
& \textbf{1.21}
& 20.15
& \underline{34.00}
& 14.48
& 10.47
\\

ISO~\cite{yu2024iso}
& \mbox{ECCV 2024}
& 20.02
& \underline{9.88}
& 0.97
& 43.79
& 2.69
& \textbf{4.41}
& \textbf{2.87}
& \underline{1.10}
& 21.25
& 31.32
& \underline{14.49}
& \underline{10.72}
\\

% ============================================================
% Ours
% ============================================================
\rowcolor{oursgray}
CrowdOcc (ours)
& 2026
& \textbf{21.12}
& \textbf{11.80}
& \underline{2.08}
& \textbf{46.23}
& \textbf{4.64}
& 3.31
& 1.41
& 0.63
& \underline{21.60}
& \textbf{37.02}
& \textbf{15.80}
& \textbf{11.40}
\\

% ============================================================
% Bottom thick rule
% ============================================================
\noalign{\vskip 1pt}
\specialrule{1.05pt}{0pt}{0pt}

\end{tabularx}
}

\vspace{3mm}

\end{table*}

\begin{figure*}[!t]
    \centering

    \includegraphics[width=\textwidth]{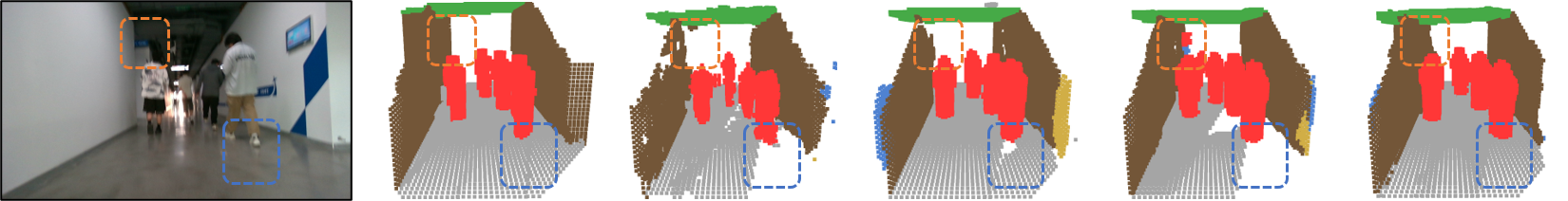}

    \vspace{-1mm}

    {\small
\noindent
\makebox[0.2270\textwidth][c]{Input}%
\makebox[0.1546\textwidth][c]{\hspace{1mm}Ground Truth}%
\makebox[0.1546\textwidth][c]{\hspace{1mm}EmbodiedOcc}%
\makebox[0.1546\textwidth][c]{\hspace{1mm}MonoScene}%
\makebox[0.1546\textwidth][c]{\hspace{1mm}ISO}%
\makebox[0.1546\textwidth][c]{\hspace{1mm}CrowdOcc (Ours)}
}

    \vspace{-5mm}

    \caption{Qualitative comparison of CrowdOcc with other monocular SSC methods.}
    
    \label{fig:qualitative_comparison}

    \vspace{1mm}
\end{figure*}

\section{EXPERIMENTS}

The acquisition platform mounts an Intel RealSense D455 RGB-D camera and a Helios-16P LiDAR on a Unitree Go1 quadruped, as shown in Fig.~\ref{fig:crowdocc_overview}. Our model is implemented in PyTorch with CUDA acceleration and trained on four NVIDIA GeForce RTX 4090-D GPUs. A frozen MoGe-2 ViT-L~\cite{wang2025moge2} provides monocular depth and surface normals. We train for 10 epochs using AdamW~\cite{loshchilov2019adamw} with a total batch size of 4 and a weight decay of \(1\times10^{-4}\). The main network and HCSI use a learning rate of \(1\times10^{-4}\), while DepthNet uses \(5\times10^{-6}\). For the normal branch, the learning rate is linearly warmed up from \(1\times10^{-5}\) to \(3\times10^{-5}\) in the first epoch and then cosine-decayed to \(5\times10^{-6}\). We set \(\lambda_h=1.0\). Unless otherwise stated, all methods follow the same CrowdOcc split, taxonomy, and evaluation protocol.

\subsection{Evaluation Metrics}

Following standard SSC protocols, we report Precision, Recall, and IoU for binary scene completion. All occupied semantic classes are merged into a single occupied category, while empty voxels form the free-space category. For semantic completion, we report the IoU of each occupied class and their mean IoU (mIoU) over the eight semantic categories. Since human occupancy is central to crowded indoor perception, Human IoU is additionally highlighted throughout the experiments. Class frequencies in Table~\ref{tab:baseline_comparison} are computed over valid occupied voxels and are reported only to characterize the semantic distribution of CrowdOcc. Invalid and ignored voxels are excluded from both scene-completion and semantic-completion metrics.

\subsection{Quantitative Comparisons}

Table~\ref{tab:baseline_comparison} reports quantitative results on the CrowdOcc test set, comprising three physical scenes disjoint from those used for training and validation. Our method achieves 15.80 IoU and 11.40 mIoU, outperforming the strongest competing results by +1.31 IoU and +0.68 mIoU, respectively. The gain is particularly pronounced for human occupancy: our method reaches 46.23 Human IoU, exceeding NDC-Scene (44.12) and ISO (43.79) by +2.11 and +2.44 points~\cite{yao2023ndcscene,yu2024iso}. It also attains the highest recall of 37.02, compared with 34.00 for NDC-Scene and 31.32 for ISO, while maintaining competitive precision. These results establish state-of-the-art SSC performance on unseen scenes, demonstrating strong cross-scene generalization.

The class-wise results show that the improvement extends beyond human occupancy. Our method achieves the best IoU on floor, wall, human, and door, reaching 21.12, 11.80, 46.23, and 4.64, respectively. In particular, wall IoU improves from 9.88 with ISO to 11.80, while floor IoU increases from 20.66 with NDC-Scene to 21.12. The consistent gains on both structural and human classes are aligned with the complementary roles of NGSGF and HCSI in geometry enhancement and human occupancy reasoning.

\subsection{Ablations}

We conduct cumulative ablations on the CrowdOcc test set in Table~\ref{tab:ablation}. Starting from the ISO baseline (Q0)~\cite{yu2024iso}, we progressively introduce MoGe-2 Depth~\cite{wang2025moge2}, HCSI, and NGSGF. In Q1, MoGe-2 replaces the original Depth Anything~\cite{yang2024depthanything} depth prior while keeping the remaining architecture unchanged, improving IoU/mIoU from 14.49/10.72 to 14.89/11.25. Adding HCSI in Q2 raises Human IoU from 42.88 to 45.77 and further improves IoU to 15.34, confirming the benefit of targeted human-centric 3D interaction.

Finally, NGSGF yields the full model (Q3), reaching 46.23 Human IoU, 15.80 IoU, and 11.40 mIoU. Compared with Q0, the complete model gains +2.44 Human IoU, +1.31 IoU, and +0.68 mIoU. The cumulative improvements are consistent with the intended roles of the three components: MoGe-2 provides a stronger depth prior, HCSI improves human occupancy reasoning, and NGSGF further enhances scene geometry through normal-guided fusion.

\subsection{Qualitative Comparisons}

Fig.~\ref{fig:qualitative_comparison} shows typical predictions on the CrowdOcc test set, where dense human presence introduces extensive human-scene occlusion and strong foreground interference. EmbodiedOcc~\cite{wu2025embodiedocc} struggles to maintain structural continuity under these incomplete observations, producing fragmented wall and floor geometry together with poor human occupancy. MonoScene~\cite{cao2022monoscene} better preserves the global layout, but the wall and floor remain discontinuous in heavily occluded regions. ISO~\cite{yu2024iso} recovers more complete scene structure, yet local floor geometry remains discontinuous and foreground human occupancy is misplaced in the background. In contrast, our method maintains coherent wall and floor geometry while preserving human occupancy at the correct spatial scenes.

\section{CONCLUSIONS}

In this paper, we introduced CrowdOcc, an RGB-D dataset for monocular SSC on quadruped robots in real crowded indoor environments. Our method targets two characteristic failures in this setting: discontinuous scene geometry under human-scene occlusion and incomplete or spatially misplaced human occupancy. NGSGF complements depth-aware lifting with surface-normal cues to improve structural geometry, while HCSI performs Top-\(K\) human-centric interaction to refine human occupancy and its spatial relation to the surrounding scene. Together, these designs improve both geometric completion and human occupancy reasoning. Experiments on test scenes disjoint from training and validation establish state-of-the-art SSC performance on CrowdOcc and demonstrate strong cross-scene generalization.

\begin{table}[!t]

% ============================================================
% Caption
% ============================================================
\refstepcounter{table}
\label{tab:ablation}

\begin{minipage}{\columnwidth}
\footnotesize
\setlength{\parindent}{0pt}
\noindent
TABLE~\thetable: Ablation study on the CrowdOcc test set. Each variant progressively adds the component indicated to the preceding configuration.
\textbf{Bold} indicates the best result and
\underline{underline} indicates the second-best result.
\end{minipage}

\vspace{5pt}

% ============================================================
% Table
% ============================================================
{
\footnotesize

\renewcommand{\arraystretch}{1.20}
\setlength{\tabcolsep}{4pt}
\arrayrulecolor{black}

\begin{tabularx}{\columnwidth}{
    >{\raggedright\arraybackslash}X
    !{\vrule width 0.4pt}
    >{\centering\arraybackslash}p{1.55cm}
    >{\centering\arraybackslash}p{1.15cm}
    >{\centering\arraybackslash}p{1.15cm}
}

% Top rule
\specialrule{1.0pt}{0pt}{0pt}
\noalign{\vskip 1.0pt}

\textbf{Variant}
& Human IoU$\uparrow$
& IoU$\uparrow$
& mIoU$\uparrow$
\\

% Middle rule
\noalign{\vskip 1.0pt}
\specialrule{0.4pt}{0pt}{0pt}
\noalign{\vskip 1.0pt}

(Q0) Baseline
& 43.79
& 14.49
& 10.72
\\

(Q1) + MoGe-2 Depth
& 42.88
& 14.89
& 11.25
\\

(Q2) + HCSI
& \underline{45.77}
& \underline{15.34}
& \underline{11.26}
\\

\rowcolor{oursgray}
(Q3) + NGSGF
& \textbf{46.23}
& \textbf{15.80}
& \textbf{11.40}
\\

% Bottom rule
\noalign{\vskip 1.0pt}
\specialrule{1.0pt}{0pt}{0pt}

\end{tabularx}
}

\vspace{-1mm}

\end{table}

\clearpage
\bibliographystyle{IEEEtran}
\bibliography{references}

\end{document}